\documentclass{article}

\usepackage{amsmath}
\usepackage{amssymb}
\usepackage{graphicx}
\usepackage{amsfonts}       % blackboard math symbols
\usepackage{nicefrac}       % compact symbols for 1/2, etc.
\usepackage{microtype}      % microtypography
\usepackage{booktabs}       % professional-quality tables
\usepackage{authblk}
\usepackage{url}

\begin{document}

% first the title is needed
\title{Neuron Segmentation Using Deep Complete Bipartite Networks\thanks{The work was supported in part by NSF grants CCF-1217906, CNS-1629914, and CCF-1617735. S. Banerjee was supported by the Department of Defense (Army Research Laboratory), 68008-LS-II. A. Grama was supported by the Cabot and Zuckerman fellowships from Harvard University.}}

\author[1]{Jianxu Chen}
\author[1]{Sreya Banerjee}
\author[2]{Abhinav Grama}
\author[1]{Walter J. Scheirer}
\author[1]{Danny Z. Chen}
\affil[1]{Department of Computer Science and Engineering, University of Notre Dame, USA}
\affil[2]{Department of Molecular and Cellular Biology, Harvard University, USA}

%%%%%%%%%%%%%%%%%%%%%%%%%%%%%%%%%%%%%%%%%%%%%%%%%%%%%%%%
%% a short form should be given in case it is too long for the running head
%\titlerunning{Lecture Notes in Computer Science: Authors' Instructions}
%\author{Alfred Hofmann%
%\thanks{Please note that the LNCS Editorial assumes that all authors have used
%the western naming convention, with given names preceding surnames. This determines
%the structure of the names in the running heads and the author index.}%
%\and Ursula Barth\and Ingrid Haas\and Frank Holzwarth\and\\
%Anna Kramer\and Leonie Kunz\and Christine Rei\ss\and\\
%Nicole Sator\and Erika Siebert-Cole\and Peter Stra\ss er}
%%
%\authorrunning{Lecture Notes in Computer Science: Authors' Instructions}
%% (feature abused for this document to repeat the title also on left hand pages)
%
%% the affiliations are given next; don't give your e-mail address
%% unless you accept that it will be published
%\institute{Springer-Verlag, Computer Science Editorial,\\
%Tiergartenstr. 17, 69121 Heidelberg, Germany\\
%\mailsa\\
%\mailsb\\
%\mailsc\\
%\url{http://www.springer.com/lncs}}
%
%%
%% NB: a more complex sample for affiliations and the mapping to the
%% corresponding authors can be found in the file "llncs.dem"
%% (search for the string "\mainmatter" where a contribution starts).
%% "llncs.dem" accompanies the document class "llncs.cls".
%%
%
%\toctitle{Lecture Notes in Computer Science}
%\tocauthor{Authors' Instructions}
%%%%%%%%%%%%%%%%%%%%%%%%%%%%%%%%%%%%%%%%%%%%%%%%%%%%%%%%
\maketitle

\begin{abstract}
In this paper, we consider the problem of automatically segmenting neuronal cells in dual-color confocal microscopy images. This problem is a key task in various quantitative analysis applications in neuroscience, such as tracing cell genesis in \textit{Danio rerio} (zebrafish) brains. Deep learning, especially using fully convolutional networks (FCN), has profoundly changed segmentation research in biomedical imaging. We face two major challenges in this problem. First, neuronal cells may form dense clusters, making it difficult to correctly identify all individual cells (even to human experts). Consequently, segmentation results of the known FCN-type models are not accurate enough. Second, pixel-wise ground truth is difficult to obtain. Only a limited amount of approximate instance-wise annotation can be collected, which makes the training of FCN models quite cumbersome. We propose a new FCN-type deep learning model, called deep complete bipartite networks (CB-Net), and a new scheme for leveraging approximate instance-wise annotation to train our pixel-wise prediction model. Evaluated using seven real datasets, our proposed new CB-Net model outperforms the state-of-the-art FCN models and produces neuron segmentation results of remarkable quality.

\end{abstract}

\section{Introduction}
\label{sec:intro}

A fundamental problem in neuroscience research is automatic image segmentation of neuronal cells, which is the basis for various quantitative analyses of neuronal structures, such as tracing cell genesis in \textit{Danio rerio} (zebrafish) brains \cite{bio_3} (e.g., using the EMD-based tracking model~\cite{EMD}). Fully convolutional networks (FCN) \cite{FCN} have emerged as a powerful deep learning model for image segmentation. In this paper, we aim to study the problem of automatically segmenting neuronal cells in dual-color confocal microscopy images with deep learning.

\begin{figure}[bt]
   \centering
   \includegraphics[width = 4.0in]{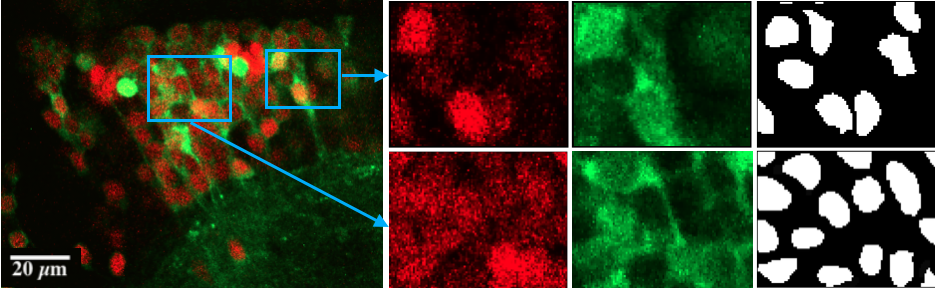} 
   \caption{A dual-color confocal microscopy image example of the tectum of a zebrafish brain. Two zoom-in regions are shown, including the red channel, the green channel, and approximate instance-wise human annotation.} 
   \label{fig:image}
\end{figure}

In this problem, we face two major challenges, which also arise in other biomedical image segmentation applications. (1) Neuron segmentation is quite complicated, due to vanishing separation among cells in densely packed clusters, very obscure cell boundaries, irregular shape deformation, etc (see Fig.~\ref{fig:image}). Even to biologists, it is difficult to correctly identify all individual cells visually. Since state-of-the-art FCN models may incur considerable errors in this difficult task, it is highly desirable to develop new effective models for it. (2) To train FCN-type models for per-pixel prediction, pixel-level supervision is commonly needed, using fully annotated images. However, in our problem, even experienced biologists can hardly determine per-pixel ground truth. For pixels near cell boundaries, even approximate ground truth is difficult to acquire. In fact, biologists only perceive instance-level information, namely, presence or absence of cells. Thus, how to leverage instance-level annotation to train pixel-level FCN models is important.

In this paper, we propose a new FCN-type segmentation model, called deep \textbf{C}omplete \textbf{B}ipartite Networks (CB-Net). Its core macro-architecture is inspired by the structure of complete bipartite graphs. Our proposed CB-Net explicitly employs multi-scale feature re-use and implicitly embeds deep supervision. Moreover, to overcome the lack of pixel-level annotation, we present a new scheme to train pixel-level deep learning models using approximate instance-wise annotation. Our essential idea is to extract reliable and discriminative samples from all pixels, based on instance-level annotation. We apply our model to segment neuronal cells in dual-color confocal microscopy images of zebrafish brains. Evaluated using 7 real datasets, our method produces high quality results, both quantitatively and qualitatively. Also, the experiments show that our CB-Net can achieve much higher precision/recall than the state-of-the-art FCN-type models.

\textbf{Related Work}. In literature, different strategies have been proposed to improve FCN-type segmentation models, most of which share some of the following three characteristics. First, FCN can be embedded into a multi-path framework, namely, applying multiple instances of FCNs through multiple paths for different sub-tasks~\cite{dcan}. An intuitive interpretation of this is to use one FCN for cell boundaries and another FCN for cell interior, and finally fuse the information from such two paths as the cell segmentation results. Second, extra pre-processing and/or post-processing can be included to boost the performance of FCNs. One may apply classic image processing techniques to the input images and combine the results thus produced together with the input images as the input to FCNs \cite{pre_processing}. Also, contextual post-processing (e.g., fully connected CRF \cite{CRF} or topology aware loss \cite{loss}) can be applied to impose spatial consistency to obtain more plausible segmentation results. Third, FCN, as a backbone network, can be combined with an object detection sub-module \cite{bottom_up_torr} or be applied in a recurrent fashion \cite{recurrent_instance} to improve instance-level segmentation accuracy. 

In this paper, we focus on developing the CB-Net model, bearing in mind that CB-Net can be viewed as a backbone network and thus be seamlessly combined with the above mentioned strategies for further improvement of segmentation.

\section{Methodology}

%We propose a new FCN model, CB-Net, whose macro-architecture has the structure of a complete bipartite graph (Section~\ref{sec:network}). Further, we propose a strategy to generate pixel-level training data from approximate instance-wise annotation (Section~\ref{sec:annotation}), aiming to make training effective even when per-pixel annotation is not available. Section~\ref{sec:training} presents our implementation and training in detail.  

\subsection{CB-Net}
\label{sec:network}

\begin{figure}[tb]
   \centering
   \includegraphics[width = 4.9in]{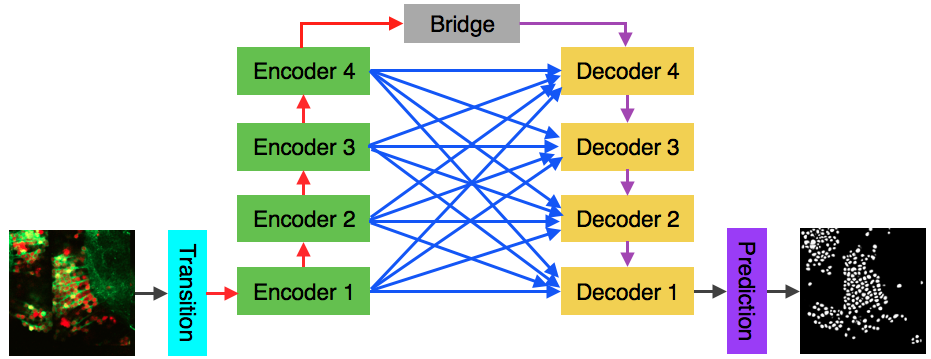} 
   \caption{The macro-architecture of CB-Net. Overall, CB-Net operates at five different scales (i.e., resolutions). The bridge block performs abstraction at the highest scale (i.e., the lowest resolution). Encoder block $k$ and decoder block $k$ process the feature space at scale $k$ ($k=1,\ldots,4$). There is a shortcut link (blue arrow) from every encoder block to every decoder block, to deeply consolidate multi-scale information.}
   \label{fig:overall}
\end{figure}

Fig.~\ref{fig:overall} shows a schematic overview of CB-Net. This model employs a generalized ``complete bipartite graph" structure to consolidate feature hierarchies at difference scales. Overall, CB-Net works at five different scales (i.e., different resolutions of the feature plane). At scale $k$ ($k=1, \ldots, 4$), an encoder block $k$ is employed to distill contextual information and a decoder block $k$ is used to aggregate the abstracted information at this scale, while the bridge block performs abstraction at the highest scale/lowest resolution (i.e., scale 5).

There is one shortcut connection between each encoder and each decoder to implement the complete bipartite structure, which implicitly integrates the benefits from diversified depths, feature reuse, and deep supervision \cite{densely}. With the interacting paths between encoder blocks and decoder blocks, the whole network implicitly ensembles a large set of sub-networks of different depths, which significantly improves the representation capacity of the network. In a forward pass, the encoded features at one scale are effectively reused to aid decoding at each scale. In a backward pass, the shortcut connections assist the gradient flow back to each encoder block efficiently, so that the supervision through the prediction block can effectively have deep impact on all encoder blocks.  

\begin{figure}[tb]
   \centering
   \includegraphics[width = 4.9in]{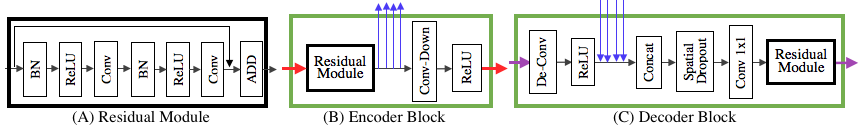} 
   \caption{The structures of the core residual module (A), an encoder block (B), and a decoder block (C). The blue arrows indicate the shortcut connections from encoder blocks to decoder blocks (better viewed in color).}
   \label{fig:encoder_decoder}
\end{figure}

 {\bf Core Blocks (Encoders and Decoders).} Fig.~\ref{fig:encoder_decoder} shows the structures of the encoder blocks and decoder blocks. A key component for feature extraction at a particular scale is the residual module \cite{identity}, with two successive ``batch normalization BN + ReLU + $3\times 3$ convolution"
 %CHEN
 % Above, why use "(BN)", not simply "BN"?  I replace (BN) by BN.
 % Also, does the component consists of "the residual module" AND "two successive ..." ?  Or, there is another interpretation of the above sentence?
 %
 (see Fig.~\ref{fig:encoder_decoder}(A)). Since we do not pad the convolution output, the input to the first BN is trimmed in both the height and width dimensions before adding to the output of the second convolution. The width of each residual module (i.e., the number of feature maps processed in the module) follows the pyramid design \cite{design_pattern}, i.e., $32k$ width at scale $k$. 
 
 The \textit{encoders} consist of a residual module and a ``Conv-Down" layer for downsampling. Inspired by \cite{all_conv}, we use a $2\!\times\!2$ convolution with stride 2, instead of pooling, to make the downsampling learnable so as to be scale-specific. The \textit{decoders} first fuse the main decoding stream with reused features from the encoders at different scales. The concatenated features include the deconvolution result \cite{FCN} from a previous decoder (or the bridge block), and 4 sets of re-sized feature maps, each from the output of a different encoder block with proper rescaling (bi-linear interpolation for up-sampling and max pooling for down-sampling) and/or border cropping . Then, a spatial dropout \cite{dropout} (the rate = 0.5), namely randomly selecting a subset of the concatenated feature maps during training, is applied to avoid overfitting to features from specific scales. Before feeding into the residual module, a $1\!\times\!1$ convolution is applied for dimension casting. % so that the number of feature maps matches the width of the residual module at the particular scale.% (i.e., $32k$ at scale $k$). 
%(Note: To concatenate features from different encoder blocks, the feature maps need to be rescaled (by bi-linear interpolation for up-sampling or max pooling for down-sampling with no learnable parameters) and/or cropped at the border to match the spatial dimension. This is because the outputs of different encoder blocks have different effective strides and no padding is applied in all residual modules.) %Table~\ref{tab:resize} summarizes the exact operations.
 
% \begin{table}[tb]
%  \caption{Re-size operation of the shortcut link from encoder block $i$ to decoder block $j$ to match the feature plane dimension. Suppose the input dual-color image patch is of size $428\times428$ pixels. The trim amount is the number of pixels cropped from each side of the feature plane (T = Trim, D = Downsampling, and U = Upsampling).}
%  \label{tab:resize}
%  \centering
%  \begin{tabular}{|c|c|c|c|c|}
%  \hline
%        & Decoder 1 & Decoder 2 & Decoder 3 & Decoder 4 \\
%    \hline
%    %\midrule
%    Encoder 1  & T(60)+D(8x) & T(76)+D(4x) & T(84)+D(2x) & T(88)     \\
%    \hline
%    Encoder 2  & T(28)+D(4x) & T(36)+D(2x) & T(40) &  U(2x)+T(84)    \\
%    \hline
%    Encoder 3  & T(12)+D(2x) & T(16) & U(2x)+T(36) &  U(4x)+T(76)     \\
%    \hline
%    Encoder 4  & T(4)  & U(2x)+T(12) & U(4x)+T(28) &  U(8x)+T(60)    \\
%    \hline
%   % \bottomrule
%  \end{tabular}
%\end{table}
 
{\bf Auxiliary Blocks.} The transition block is a $7\!\times\!7$ convolution and ReLU (with zero padding), which can be interpreted as a mapping from the input space (of dimension 2, red/green channels, in our case) to a rich feature space~\cite{design_pattern} for the model to exercise its representation power. The bridge block, similar to encoders but no down-sampling, aims to perform the highest level abstraction and trigger the decoding stream. The prediction block is a $1\!\times\!1$ convolution and LogSoftMax, whose output indicates the probability of each pixel belonging to a neuron. 

\subsection{Leveraging Approximate Instance-wise Annotation}
\label{sec:annotation}

\begin{figure}[tb]
   \centering
   \includegraphics[width = 4.9in]{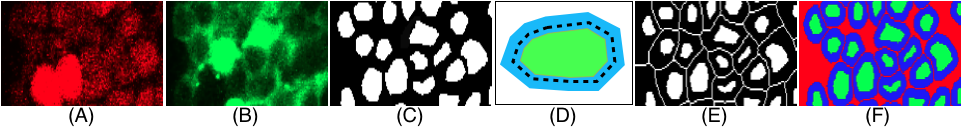} 
   \caption{Illustration of pixel-wise training data generation from approximate instance-wise annotation made by human. (A-B) Input red/ green fluorescent channels. (C) Approximate instance-wise human annotation. (D) Illustrating the ``fuzzy boundary" of a cell, i.e., the blue ring area. The dotted curve is the boundary of the human-annotated region. The pixels in the green area are assigned a label of ``Cell". (E) Illustrating the outer medial axis. (F) The generated pixel-wise training data: green=``Cell", red=``Background", and blue=``Fuzzy Boundary" (better viewed in color).}
   \label{fig:annotation}
\end{figure}

In our problem, per-pixel ground truth cannot be obtained, even by experienced biologists. Instead, human experts are asked to draw a solid shape within each cell to indicate the cell body approximately. (Note: By ``approximate", we mean that we know neither the exact bounding box nor the exact shape of each instance.) Generally, the annotations are drawn in a conservative manner, namely, leaving uncertain pixels close to cell boundaries as unannotated. But, when it is absolutely sure, the sizes of the solid shapes are drawn as large as possible. In Fig.~\ref{fig:annotation}(C), all annotated regions are in white, and the remaining pixels are in black. Directly using this kind of annotation as per-pixel ground truth will cause considerably many positive samples (i.e., pixels of cells) being used falsely as negative samples (i.e., background), due to such conservative annotation. 

Our main idea of utilizing approximate instance-wise annotation for pixel-level supervision is to extract a sufficient number of more reliable and more effective samples from all pixels based on the available annotations. Specifically, (1) we prune the annotated regions to extract reliable ground truth pixels belonging to cells, and (2) we identify a subset of all unannotated pixels that is more likely to be background, especially in the gap areas among touching cells.

Let $A$ be an annotated binary image. First, we perform erosion on $A$ (with a disk template of radius 1); let $E$ be the resulting eroded regions. Second, we perform dilation on $A$ (with a disk template of radius 4); let $D$ be the result. Third, we compute the outer medial axis of $E$ (see Fig.~\ref{fig:annotation}(E)), denoted by $M$. Then, for each pixel $p$, we assign its label $L(p)$ as: 1 (Cell), if $p\in E$; 2 (Background), if $p\in M\!\cup\!(A\!\setminus\!D)$; 3 (Fuzzy Boundary), otherwise. The ``Fuzzy Boundary" (roughly a ring along the boundary of an annotated region, see Fig.~\ref{fig:annotation}(D)), where the pixel labels are the most uncertain, will be ignored during training. A special scenario is that such ring shapes for proximal cells may overlap. So, the outer medial axis of the eroded annotated regions is computed and is retained as the most representative background samples to ensure separation. Note that this scheme may also be applied to other applications by adjusting the parameters (e.g., larger erosion for less conservative annotation).

\subsection{Implementation Details}
\label{sec:training}

\textbf{Post-processing.} The output of CB-Net can be viewed as a probability map, in which each pixel is given a probability of being in a cell (a value between 0 and 1). We produce the final binary segmentation by thresholding (at 0.75), two successive binary openings (with a disk template of radius 5, and a square template of size 3), and hole filling. We find the CB-Net prediction is of high accuracy so that the threshold is not sensitive and simple morphological operations are sufficient to break the potentially tenuous connections among tightly touching cells (not common, less than $5\%$). Also, the template sizes of the morphological operations are determined based on our object shapes (i.e., cells), and should not be difficult to adjust for other applications (e.g., a larger template for larger round cells, or a smaller template for star shape cells with tenuous long ``arms").

\textbf{Data Augmentation.} Since we have only 5 images with annotation, we perform intensive random data augmentation to make effective training and reduce overfitting. In each iteration, an image patch is processed by (1) horizontal flip, (2) rotation by a random degree (an integer between 1 and 180), or (3) vertical flip. Each flip is randomly applied with a probability of $50\%$. Because the random rotation usually involves intensity interpolation, implicitly introducing lighting noise, no color jittering is employed. 

\textbf{Training.} Learnable parameters are initialized as in \cite{initialization} and optimized using Adam scheme \cite{adam}. The key hyperparameters are determined empirically: (1) We use batch size of 1, since large image patch is preferred over large batch size \cite{unet}. (2) We use higher learning rates for a few epochs (1e-5 for epochs 1-50 and 1e-6 for epochs 51-100), and fix a small learning rate, 1e-7, for all the remaining epochs. (3) We use a weighted negative log likelihood criterion (0.25, 0.75, and 0 for the ``Cell", ``Background", and ``Fuzzy Boundary" weights, respectively). Thus, the fuzzy boundary is ignored by assigning a zero weight. The background is associated with a higher weight to encourage separation among cells.  %This is because we encourage to assign "background" over "cell" within the uncertain region, i.e., the fuzzy boundary. In other words, we intentionally encourage separation between cells.

% \begin{figure}[tb]
%   \centering
%   \includegraphics[width = 4.24in]{./fig/snake.png} 
%   \caption{Sample results by a classic image segmentation method and CB-Net. Yellow arrows indicate several regions (not all) where considerable false negatives occur.}
%   \label{fig:snake}
%\end{figure}

 \begin{figure}[tb]
   \centering
   \includegraphics[width = 4.9in]{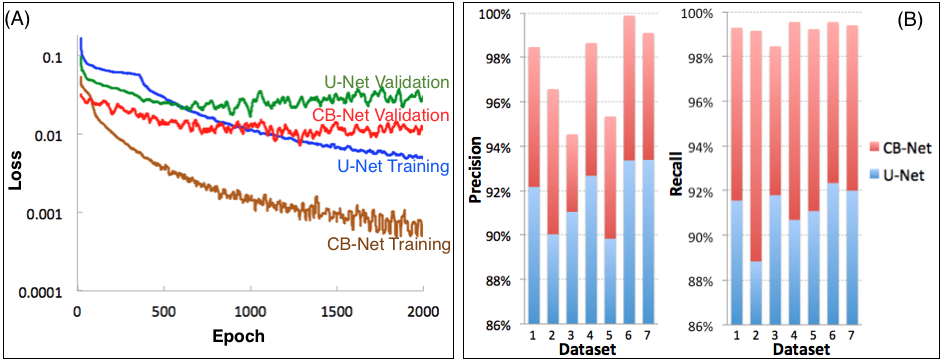} 
   \caption{(A) The results of the leave-one-out experiments; (B) the precision and recall of U-Net and CB-Net on seven different real datasets.}
   \label{fig:result}
\end{figure}

\section{Experiments}
\label{sec:exp}

%\subsection{Applications and Dataset Collection}
%\label{sec:app}

%Larval zebrafish is a popular model system for investigating neural circuits underlying visual computations and behaviors because of its translucent body, well established genetics, and robust visual behaviours~\cite{bio_neural}. The primary target, the tectum, continues to increase in size as the fish grows, in the form of the addition of newborn neurons to pre-existing circuits \cite{bio_3}. Thus, the larval zebrafish tectum presents itself as an accessible system to study the mechanism of stable integration of new cells in a functioning network. This could prove to be useful in further research on how to repair damaged brain tissues.

Besides having 5 images for training, we use 7 in-house datasets for evaluation, each containing 55 dual-color microscopy images of a zebrafish brain. We use double transgenic fish where GCaMP6s, a green fluorescent protein (GFP) based genetically encoded calcium indicator, and H2b-RFP, a histone fused red fluorescent protein (RFP), are driven by the elavl3 promoter. This yields dual-color images, in which all neurons in the double transgenic fish express green fluorescence in the cytosolic compartment and red fluorescence in the nucleus. 

Our method is compared with U-Net \cite{unet}, a state-of-the-art FCN-type model, which has achieved lots of successes in various biomedical image segmentation applications. For fair comparison, we use the same training procedure to train U-Net as we do for CB-Net. The numbers of learnable parameters for CB-Net and U-Net are 9M and 31M, respectively. Due to the multi-scale feature reuse, a smaller width is sufficient for each residual module in CB-Net. Consequently, CB-Net contains $70\%$ fewer learnable parameters than U-Net.

\textit{Leave-one-out experiments} are conducted to quantitatively assess the performance. The results of running 2000 training epochs are given in Fig.~\ref{fig:result}(A). One can observe that CB-Net can achieve better validation performance than U-Net, and overfitting is not a severe issue even using only 5 annotated training images.

 \begin{figure}[tb]
   \centering
   \includegraphics[width = 3in]{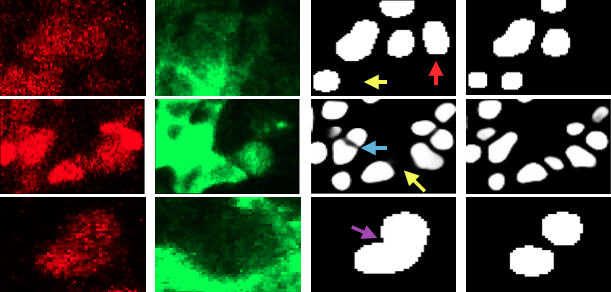} 
   \caption{Sample results of U-Net and our CB-Net. Left to right: Red and green fluorescence channels, results of U-Net and CB-Net. Some errors are indicated by arrows: yellow (false negative), red (false positive), blue (false split), and purple (false merge).}
   \label{fig:quality}
\end{figure}

\textit{Performance on the real datasets} was examined in a proof-reading manner. This is because pixel-level ground truth is not available in our problem (see Section~\ref{sec:intro}), and even approximate instance-level annotation can take two experts over 20 hours in total to manually annotate 5 images for training. Strictly speaking, we presented the segmentation results to experienced biologists in order to (1) confirm true positives, (2) reject false detections, and (3) detect false negatives. Note that falsely merged or falsely separated cells are treated as false detection. If a segmented cell is much smaller (resp., larger) than the actual size, then it is classified as false negative (resp., false detection). Finally, Precision and Recall are calculated. In fact, the proof-reading evaluation for our problem is too time consuming to make extensive quantitative ablation evaluation in practice. Also, with a similar amount of effort, we choose to evaluate and compare with the most representative baseline models on many different datasets, instead of comparing with more baseline models on only few datasets. The quantitative testing results are shown in Fig.~\ref{fig:result}(B), and qualitative results are presented in Fig.~\ref{fig:quality}. It is clear that our CB-Net achieves much better results than U-Net. 

We observe that a large portion of errors made by U-Net occurs in the following two situations: (1) confusion between noisy areas and cells with relatively weak fluorescent signals (see row 1 in Fig.~\ref{fig:quality}), and (2) confusion between touching cells and large single cells (see rows 2 and 3 in Fig.~\ref{fig:quality}). The higher representative capability of CB-Net (due to the complete bipartite graph structure) enables it to extract features more effectively and gain deeper knowledge of the semantic context. Consequently, CB-Net can attain more accurate segmentation in the above two difficult situations and achieve significant improvement over U-Net.

\section{Conclusions}

In this paper, we proposed a new FCN model, CB-Net, for biomedical image segmentation. The main advantage of CB-Net is deep multi-scale feature reuse by employing a complete bipartite graph structure. Moreover, we presented a new scheme for training a pixel-wise prediction model using only approximate instance-wise annotation. Qualitative and quantitative experimental results show that our new method achieves high quality performance in automatic segmentation of neuron cells and outperforms U-Net, a state-of-the-art FCN model.

\bibliographystyle{abbrv}
\bibliography{main}

\begin{thebibliography}{10}

\bibitem{bottom_up_torr}
A.~Arnab and P.~Torr.
\newblock Bottom-up instance segmentation using deep higher-order {CRFs}.
\newblock {\em arXiv preprint arXiv:1609.02583}, 2016.

\bibitem{loss}
A.~BenTaieb and G.~Hamarneh.
\newblock Topology aware fully convolutional networks for histology gland
  segmentation.
\newblock In {\em MICCAI}, pages 460--468, 2016.

\bibitem{bio_3}
K.~L. Cerveny, M.~Varga, and S.~W. Wilson.
\newblock Continued growth and circuit building in the anamniote visual system.
\newblock {\em Developmental Neurobiology}, 72(3):328--345, 2012.

\bibitem{dcan}
H.~Chen, X.~Qi, L.~Yu, and P.-A. Heng.
\newblock {DCAN}: Deep contour-aware networks for accurate gland segmentation.
\newblock {\em arXiv preprint arXiv:1604.02677}, 2016.

\bibitem{EMD}
J.~Chen, C.~W. Harvey, M.~Alber, and D.~Z. Chen.
\newblock A matching model based on earth mover’s distance for tracking
  {M}yxococcus xanthus.
\newblock In {\em MICCAI}, pages 113--120, 2014.

\bibitem{CRF}
L.-C. Chen, G.~Papandreou, I.~Kokkinos, K.~Murphy, and A.~L. Yuille.
\newblock Semantic image segmentation with deep convolutional nets and fully
  connected {CRFs}.
\newblock {\em arXiv preprint arXiv:1412.7062}, 2014.

\bibitem{initialization}
K.~He, X.~Zhang, S.~Ren, and J.~Sun.
\newblock Delving deep into rectifiers: Surpassing human-level performance on
  {ImageNet} classification.
\newblock In {\em CVPR}, pages 1026--1034, 2015.

\bibitem{identity}
K.~He, X.~Zhang, S.~Ren, and J.~Sun.
\newblock Identity mappings in deep residual networks.
\newblock {\em arXiv preprint arXiv:1603.05027}, 2016.

\bibitem{densely}
G.~Huang, Z.~Liu, and K.~Q. Weinberger.
\newblock Densely connected convolutional networks.
\newblock {\em arXiv preprint arXiv:1608.06993}, 2016.

\bibitem{adam}
D.~Kingma and J.~Ba.
\newblock Adam: A method for stochastic optimization.
\newblock {\em arXiv preprint arXiv:1412.6980}, 2014.

\bibitem{FCN}
J.~Long, E.~Shelhamer, and T.~Darrell.
\newblock Fully convolutional networks for semantic segmentation.
\newblock In {\em CVPR}, pages 3431--3440, 2015.

\bibitem{recurrent_instance}
B.~Romera-Paredes and P.~Torr.
\newblock Recurrent instance segmentation.
\newblock {\em arXiv preprint arXiv:1511.08250}, 2015.

\bibitem{unet}
O.~Ronneberger, P.~Fischer, and T.~Brox.
\newblock {U-Net}: Convolutional networks for biomedical image segmentation.
\newblock In {\em MICCAI}, pages 234--241, 2015.

\bibitem{pre_processing}
S.~K. Sadanandan, P.~Ranefall, and C.~W{\"a}hlby.
\newblock Feature augmented deep neural networks for segmentation of cells.
\newblock In {\em ECCV}, pages 231--243, 2016.

\bibitem{design_pattern}
L.~N. Smith and N.~Topin.
\newblock Deep convolutional neural network design patterns.
\newblock {\em arXiv preprint arXiv:1611.00847}, 2016.

\bibitem{all_conv}
J.~T. Springenberg, A.~Dosovitskiy, T.~Brox, and M.~Riedmiller.
\newblock Striving for simplicity: The all convolutional net.
\newblock {\em arXiv preprint arXiv:1412.6806}, 2014.

\bibitem{dropout}
J.~Tompson, R.~Goroshin, A.~Jain, Y.~LeCun, and C.~Bregler.
\newblock Efficient object localization using convolutional networks.
\newblock In {\em CVPR}, pages 648--656, 2015.

\end{thebibliography}

\end{document}